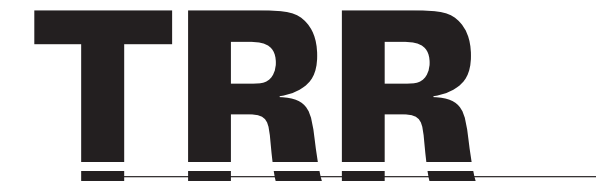

Research Article

# Intelligent Anomaly Detection for Lane Rendering Using Transformer with Self-Supervised Pretraining and Customized Fine-Tuning

**Yongqi Dong[1,2*], Xingmin Lu[3*], Ruohan Li[4], Wei Song[5], Bart van Arem[1], and Haneen Farah[1]**

**Abstract**

The burgeoning navigation services using digital maps provide great convenience to drivers. Nevertheless, the presence of anomalies in lane-rendering map images occasionally introduces potential hazards, as such anomalies can mislead human drivers and consequently contribute to unsafe driving. In response to this concern to accurately and effectively detect the anomalies, this paper transforms lane-rendering image anomaly detection into a classification problem and proposes a four-phase pipeline: data preprocessing, self-supervised pretraining with the masked image modeling (MiM) method, customized fine-tuning using cross-entropy-based loss with label smoothing, and post-processing. Leveraging state-of-the-art deep learning techniques, especially those involving transformer models, the pipeline demonstrates superior performance verified through various experiments. Notably, self-supervised pretraining with MiM can greatly enhance detection accuracy while significantly reducing the total training time. For instance, employing the Swin Transformer with Uniform Masking as self-supervised pretraining yielded a higher accuracy of 94.77% and an improved area under the curve (AUC) score of 0.9743 compared with the pure Swin Transformer without pretraining with an accuracy of 94.01% and an AUC of 0.9498. Furthermore, fine-tuning epochs were dramatically reduced to 41 from the original 280. Ablation study with regard to techniques to alleviate the data imbalance between normal and abnormal instances further reinforces the model's overall performance. In conclusion, the proposed pipeline, with its incorporation of self-supervised pretraining using MiM and other advanced deep learning techniques, emerges as a robust solution for enhancing the accuracy and efficiency of lane-rendering image anomaly detection in digital navigation systems.

**Keywords**

anomaly detection, lane rendering image, Transformer, masked image modeling, self-supervised learning, image reconstruction, image classification

With the increase in private car ownership and the emergence of information and communication technology, navigation services have become popular, gaining increasing importance, forming a crucial component in driving, and providing convenience for drivers. Navigation services are always backed up by digital map applications (*1*, *2*). A critical aspect of digital maps is the background, which is generated through data rendering. However, lane-level rendered map images may contain anomalies (errors, defects, or both), such as irregular shapes and missing edges or corners. Examples of such anomalies are shown in Figure 1. These anomalies can be confusing for drivers, impairing their understanding

[1]Faculty of Civil Engineering and Geosciences, Delft University of Technology, Delft, The Netherlands
[2]Institute of Highway Engineering, RWTH Aachen University, Aachen, Germany
[3]School of Electrical and Control Engineering, North China University of Technology, Beijing, P.R. of China
[4]Department of Civil and Environmental Engineering, College of Engineering, Villanova University, Villanova, PA
[5]School of Information Science and Technology, North China University of Technology, Beijing, P.R. of China

[*]These authors contributed equally to this work and should be considered as co-first authors.

**Corresponding Authors:**
Yongqi Dong, yongqi.dong@rwth-aachen.de
Wei Song, songwei@ncut.edu.cn

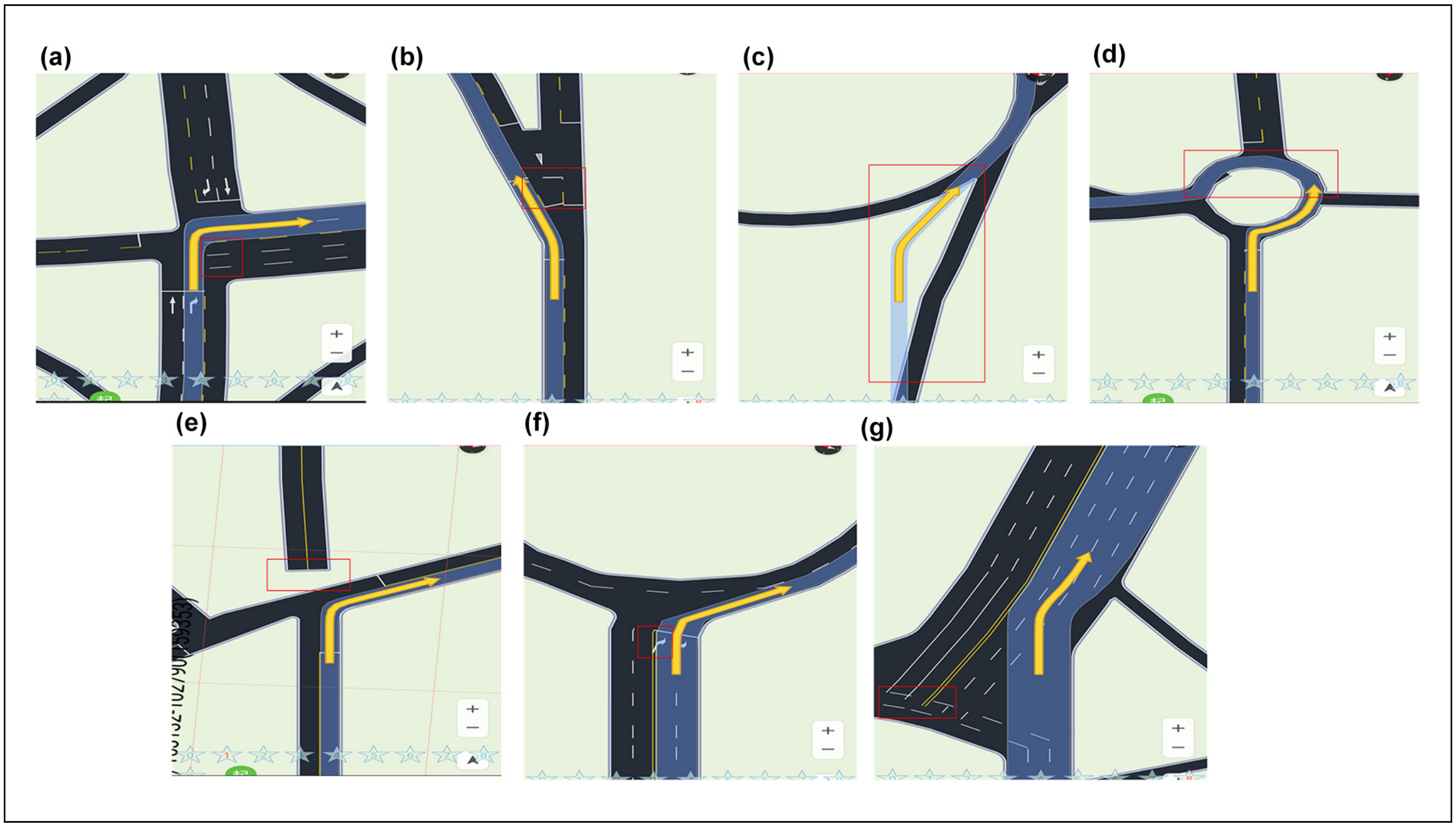

**Figure 1.** Examples of anomalous lane-rendering images. (*a*) Anomaly_1: The road center line extends out of the junction. (*b*) Anomaly_2: The stop line is in the middle of a road. (*c*) Anomaly_3: The navigation route does not match actual roads. (*d*) Anomaly_4: The road shoulder is bumpy. (e) Anomaly_5: A part of the road is missing. (*f*) Anomaly_6: The road marking arrows overlap. (*g*) Anomaly_7: The lane lines overlap. The red boxes mark the specific regions where the anomalies are.

and decision-making during navigation, which might result in critical unsafe situations.

Similar anomalies can occur in high-definition (HD) maps used by automated vehicles (AVs) (*3*, *4*). Accurate lane rendering in such maps is essential for various systems, including automated driving systems, advanced driver-assistance systems (ADAS), and smart traffic management systems, all of which rely heavily on precise and reliable mapping data to function effectively and safely. Anomalies in such maps can lead AVs into unsafe regions or induce dangerous driving behaviors.

Furthermore, this targeted problem is closely related to and can be easily transformed into relevant critical and practical real-world applications, such as road anomaly detection (*5*, *6*), road defect detection (*7*, *8*), as well as anomaly detection for lane and pavement marking on roads (*9–11*). These issues are even more crucial for road safety. It has been found that lane-related errors contribute to more than 10% of lane-change crashes (*12*), and misperception of lanes or lane boundaries is a leading factor in AV disengagements (*13*, *14*). Thus, for example, the Federal Highway Administration in the USA has detailed guidelines on pavement markings essential for safe navigation and traffic management (*15*). Similarly, China's Ministry of Transport emphasizes the importance of accurate lane marking for reducing accidents and enhancing road safety (*16*).

Overall, it is vital to correctly detect these anomalies to prevent such unsafe situations. Fortunately, with the advancement of artificial intelligence algorithms, particularly in the domain of computer vision, it is now possible to carry out intelligent and automatic anomaly detection.

Conventional studies with regard to anomaly detection in the relevant transportation domains principally focus on road-surface anomalies (*5*, *17*), road-traffic anomalies (*18*, *19*), in-vehicle and vehicle-to-vehicle communication anomalies (*20*, *21*), abnormal driving behaviors (*22–24*), and so forth. Multimodal and multisource data have been utilized with various machine learning methods to do the detection. However, few studies have employed self-supervised methods to leverage unlabeled data. On the other hand, masked autoencoders and, more generally, masked image modeling (MiM) have become popular pretraining paradigms for self-supervised visual representation learning tasks. In MiM, a portion (usually a high ratio of $\geqslant 50\%$) of the input image is randomly masked using patches, and the model tries to reconstruct the masked pixels according to the target representations. The pretrained model weights through MiM can be transferred to the downstream task for fine-tuning. Evidence in recent studies (e.g., Bao et al. [*25*], El-Nouby et al. [*26*], He et al. [*27*], Xie et al. [*28*], and Li and Dong [*29*]) has demonstrated that self-supervised pretraining with MiM can boost the downstream tasks (e.g., classification, segmentation, and object detection) to achieve more desirable performance. Thus, it is worth exploring MiM-based pretraining for anomaly detection.

Furthermore, although various image datasets (e.g., animals, digital numbers, industrial inspection image MVTec AD datasets [*30*]) and vision-based anomaly detection methods have been developed (*31–35*), to the best of the authors' knowledge and after extensive review, there are no studies that tackle abnormal lane-rendering images in digital navigation maps.

To fill the aforementioned research gaps, this study develops a four-phase pipeline with self-supervised pretraining and customized fine-tuning and uses state-of-the-art transformer models (*25*, *36–40*) to accurately and effectively detect lane-rendering image anomalies. A large-scale lane-rendering image dataset adjusted from the 2022 Global AI Challenge with both labeled and unlabeled data was adopted, and extensive experiments were carried out tackling the lane-rendering image anomaly detection problem as a two-, eight-, or nine-class classification task. Two MiM-based self-supervised pretraining methods (i.e., Uniform Masking [*39*] and Bidirectional Encoder representation from Image Transformers [BEiT] [*25*]) were customized and implemented. Extensive experiments, including ablation studies and comparative benchmarking, validate the pipeline's efficacy. To summarize, the main contributions of this paper lie in:

1. Problem reformulation: Transforming the lane-rendering anomaly detection problem into a two-, eight-, or nine-class classification problem.
2. Optimized pipeline: Proposing a four-phase pipeline with specially self-supervised pretraining and customized fine-tuning to tackle the lane-rendering image anomaly detection problem.
3. Utilization and implementation of MiM methods: Customizing and implementing two MiM self-supervised pretraining methods within the proposed four-phase pipeline. Extensive training, fine-tuning, and validating experiments demonstrated that, with MiM, the detection performance was greatly enhanced with improved area under the curve (AUC) and reduced fine-tuning epochs.
4. State-of-the-art performance: Under the proposed pipeline, the best model delivered a performance with an accuracy of 94.82%, an AUC of 0.9756, and an F1 score of 0.7879, outperforming baseline models (e.g., Vision Transformer [ViT] [*40*] and Swin Transformer [*37*]).

Note that the methods and models developed in this study can not only effectively detect lane-rendering image anomalies but also be readily adapted for related applications, such as detecting road-surface anomalies and identifying abnormal lane markings.

The rest of this paper is arranged as follows. The next section describes the research methodology, consisting of the proposed pipeline in detail, including the overall framework, data preprocessing, self-supervised pretraining, customized fine-tuning, and post-processing. Following this, the experimental set-up and results are described, comparing different models within the proposed pipeline. Then, the next section introduces methods to alleviate data imbalance. The final section summarizes the findings, lists the limitations, and proposes insights for future studies.

## Methodology

In this section, the proposed method is introduced in detail. First, the overall architecture of the proposed four-phase pipeline is illustrated and briefly explained. Then, each of the four phases (i.e., image preprocessing, self-supervised pretraining, fine-tuning classification, and post-processing) is depicted with comprehensive delineations sequentially.

### *Overall Pipeline Description*

This study proposes a pipeline of four phases to tackle the anomaly detection task for lane-rendering images in digital navigation applications. The overall pipeline of the four-phase method is illustrated in Figure 2.

The designed four phases are: (1) image preprocessing, which normalizes the inconsistent images into a uniform format, size, and resolution; (2) self-supervised pretraining, which is tackled by the MiM method using mean square error (MSE) loss and outputs the pretrained model; (3) customized fine-tuning, which adopts the pretrained model weights and further fine-tunes the neural network model as a classification task using cross-entropy-based loss (or its variants) with label smoothing; and (4) post-processing, which transforms the results of the last neural network layer (i.e., the output layer) into classification probabilities and outputs the final detection results with a tuned probability threshold. The following subsections explain these four phases in more detail.

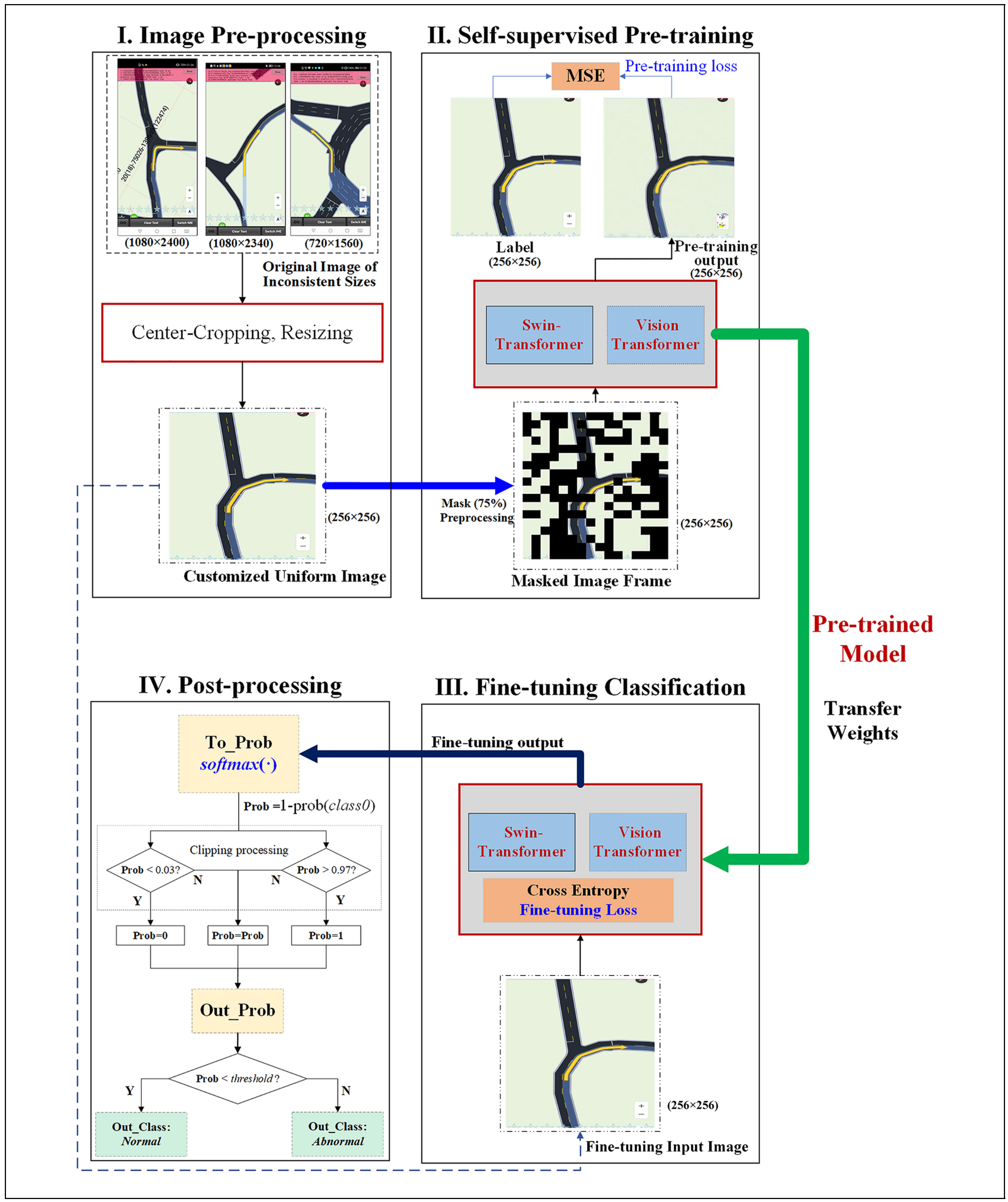

**Figure 2.** Architecture of the proposed four-phase pipeline.
*Note*: Class 0 is the normal class. MSE = mean square error.

## Image Preprocessing

This study adopts the large-scale lane-rendering image dataset adjusted and rearranged from the 2022 Global AI Challenge. The original images provided are of different resolutions and sizes. The majority of them have a resolution of 1,080 × 2,400, whereas there are a few images with different resolutions (i.e., 1,080 × 2,340 and 720 × 1,560). Furthermore, to focus on the relevant content of the images, the study identifies that the top and bottom portions contain non-map-related regions. Therefore, this study first carried out a center-cropping operation by removing the 1,080 × 300 pixels at the top and the 1,080 × 240 pixels at the bottom of the images. Then, the images were scaled to the same resolution (i.e., 256 × 256). Moreover, since the images are only partly labeled with ground truth (i.e., class label of normal or anomaly type) and a large proportion of the images are unlabeled, this study constructed a pretraining dataset with both labeled images and unlabeled images, a fine-tuning dataset with a randomly selected partly labeled image, and a testing dataset with a small proportion of the labeled images that was not seen in the fine-tuning dataset.

Similar image datasets can be created for other navigation maps by taking screenshots of the application software interface and applying the aforementioned preprocessing steps. The same process can be applied to real-world image datasets collected by cameras for anomaly detection of, for example, road lane-line markings or pavement markings. It is important that after the image preprocessing phase, the images are in a uniform format, size, and resolution.

## Self-Supervised Pretraining

For the lane-rendering images in the navigation map applications, lane lines account for only a small fraction of the whole image, as shown in Figure 1. There are seven types of anomaly in the studied dataset, whereas the majority of the lane-rendering images are normal ones. In these circumstances, it is assumed that there is more spatial redundancy with regard to image features for the abnormal lane-rendering image detection task, and thus stronger feature extraction ability is required. Therefore, it is necessary to design a method to fully extract aggregated context information as well as the critical features and correlations among pixels. Furthermore, as the examined dataset consists of many unlabeled images (>80%), it is also vital to establish a pipeline to make full use of these unlabeled images.

Motivated by the aforementioned issues, this study proposes and customizes the MiM method for self-supervised pretraining. In this phase, the total set of images serves as the input for model pretraining, regardless of whether labeled or unlabeled. The input image is randomly masked using patches, and the pretraining model tries to reconstruct the masked pixels to match the target original images. Generally, the standard objective of self-supervised pretraining with MiM can be mathematically represented by Equation 1:

$$\min \frac{1}{\Omega(\boldsymbol{i}_M)} ||\mathbf{r}_M - \mathbf{i}_M||_2 \tag{1}$$

where $\mathbf{i}, \mathbf{r} \in \mathbb{R}^{3 \times H \times W}$ are the input original red-green-blue (RGB) values and the reconstructed RGB values, respectively; $H$ is the height of the image and $W$ is the width of the image (with $H \times W = 256 \times 256$ in this study); $M$ represents the set of masked image pixels; $\Omega(\cdot)$ is the cardinality operator function to obtain the number of elements; and $||\cdot||_2$ stands for the $\ell_2$ norm. Accordingly, the objective involves minimizing the root mean squared error, $\ell_2$ loss, between the original and reconstructed pixel values for the masked regions. By focusing on accurately reconstructing the masked regions, the MiM approach encourages the model to learn rich and context-aware representations of the input image, which are crucial for downstream tasks.

Generally, there are two styles of implementing MiM: (1) raw pixel value regression, where the model directly reconstructs pixel values, and (2) converting the masked pixel signals into clusters or classes through methods such as vision tokenization (*25*, *41*) or color clustering (*42*) followed by performing a classification task for masked image prediction. Accordingly, this paper customizes and implements two distinct MiM methods: Uniform Masking (*39*) and the method introduced in BEiT (*25*). The Uniform Masking method was selected because it successfully enables efficient asymmetric structure, likewise in He et al. (*27*), of pixel-based masked autoencoder (MAE)-style self-supervised pretraining, particularly for Pyramid-based ViTs. On the other hand, BEiT was selected because it serves as a typical and well-established representation of token-based methods. BEiT is the first to successfully adapt Masked Language Modeling techniques from the Natural Language Processing (NLP) domain to the computer vision domain using ViT models. By introducing a discrete tokenization mechanism for MiM, BEiT enables ViTs to process images in a manner analogous to how transformers handle textual data, marking a significant milestone in bridging the gap between NLP and computer vision tasks.

With regard to the Uniform Masking method, two key operations play a central role in the self-supervised learning process:

1. *Uniform sampling*: This step ensures that one random patch is sampled from each 2 × 2 grid of

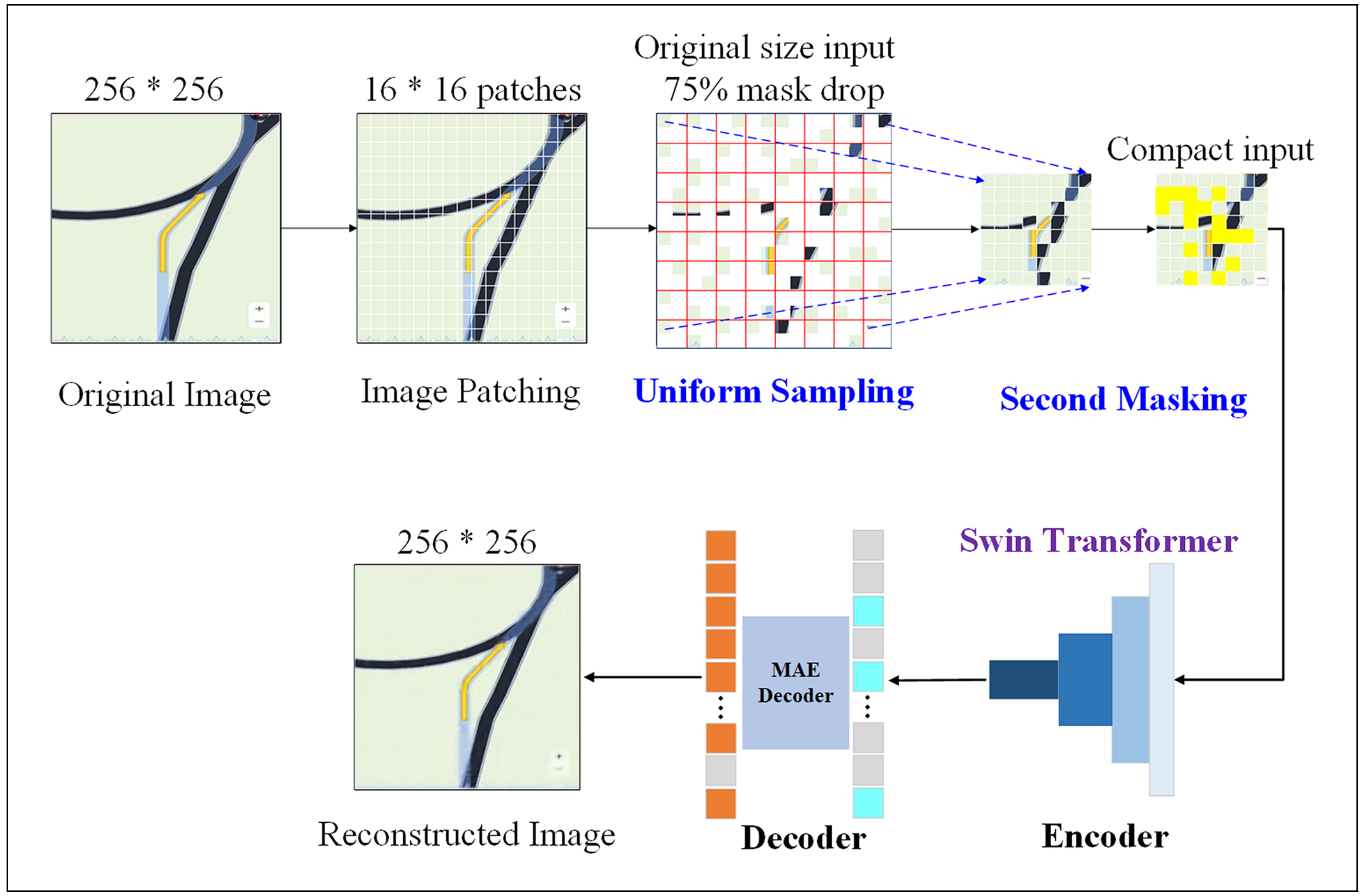

**Figure 3.** Uniform Masking method pipeline for masked image modeling.

patches within the image. As a result, 75% of the targeted region is dropped, which enforces a uniform yet sparse sampling pattern across the image.

2. *Secondary masking*: Since using only the uniform sampling can potentially make the self-supervisory task less challenging and largely hinders the representation quality (*27*), after uniform sampling, an additional random masking operation (termed "secondary masking") is applied to the sampled regions, further masking 25% of them (as used in this study) as shared learnable tokens.

Integrating uniform sampling and secondary masking together enables the pretraining method to support Pyramid-based ViTs (e.g., Liu et al. [*37*] and Wang et al. [*43*]) while preserving more transferable visual representations. The Uniform Masking method pipeline for self-supervised learning is illustrated in Figure 3. The image is first divided into 16 × 16 patches for Uniform Sampling, which drops up to 75% of the original image, and the secondary masking is operated on the remaining patches. A compact two-dimensional input, reduced to a quarter of the original image size, is constructed using the uniform-sampled patches combined with the secondary-masked tokens and is subsequently fed to the encoder. For the Pyramid-based ViT encoder, this study employs the Swin Transformer (*37*), which leverages a hierarchical architecture to effectively capture both local and global features, ensuring robust feature representation. For the decoder, the lightweight MAE Decoder, based on Vanilla ViT, is utilized, as adopted in He et al. (*27*). The MAE Decoder reconstructs the image using the encoder output features into the original size. These combinations ensure an efficient and effective architecture for self-supervised learning.

The selection of the masked ratio at 75% in the uniform sampling process is based on the experiment results reported in He et al. (*27*) and Li et al. (*39*), whereas the selection of the secondary masking ratio of 25% is based on the ablation experiment results reported in Li et al. (*39*).

With regard to the BEiT self-supervised MiM method in Bao et al. (*25*), each image is pretrained with two complementary views: image patches (e.g., 16 × 16 pixels) and visual tokens (i.e., discrete tokens). Figure 4 illustrates the method pipeline of BEiT for self-supervised MiM learning. The images are first "tokenized" into discrete visual tokens, which correspond to indices within a learned visual vocabulary. In this study, the visual vocabulary is generated using a discrete variational autoencoder (dVAE) tokenizer, as in Bao et al. (*25*) and Ramesh et al. (*41*). Following tokenization, some image patches are randomly masked and replaced with a special mask embedding before being fed into the ViT backboned encoder. Then, the objective of the self-supervised MiM pretraining task involves predicting the visual tokens of the original image from the encoded representations of the corrupted image, which effectively enables the model to learn robust visual features. The prediction of the visual tokens is handled by the MiM head, which consists of a single linear layer that converts the encoded features from the ViT encoder into a format compatible with the visual token space. Since the task involves finding the correct classes (i.e., the visual token indices), the cross-entropy loss function is employed for optimization. To reconstruct the full image, the dVAE decoder takes the predicted discrete tokens as input and reconstructs their corresponding image patches. It is important to note that the MiM head is only used during the pretraining phase; during fine-tuning, task-specific decoders replace the MiM head. In this study, the original fine-tuned hyperparameters and network architecture from Bao et al. (*25*) are adopted.

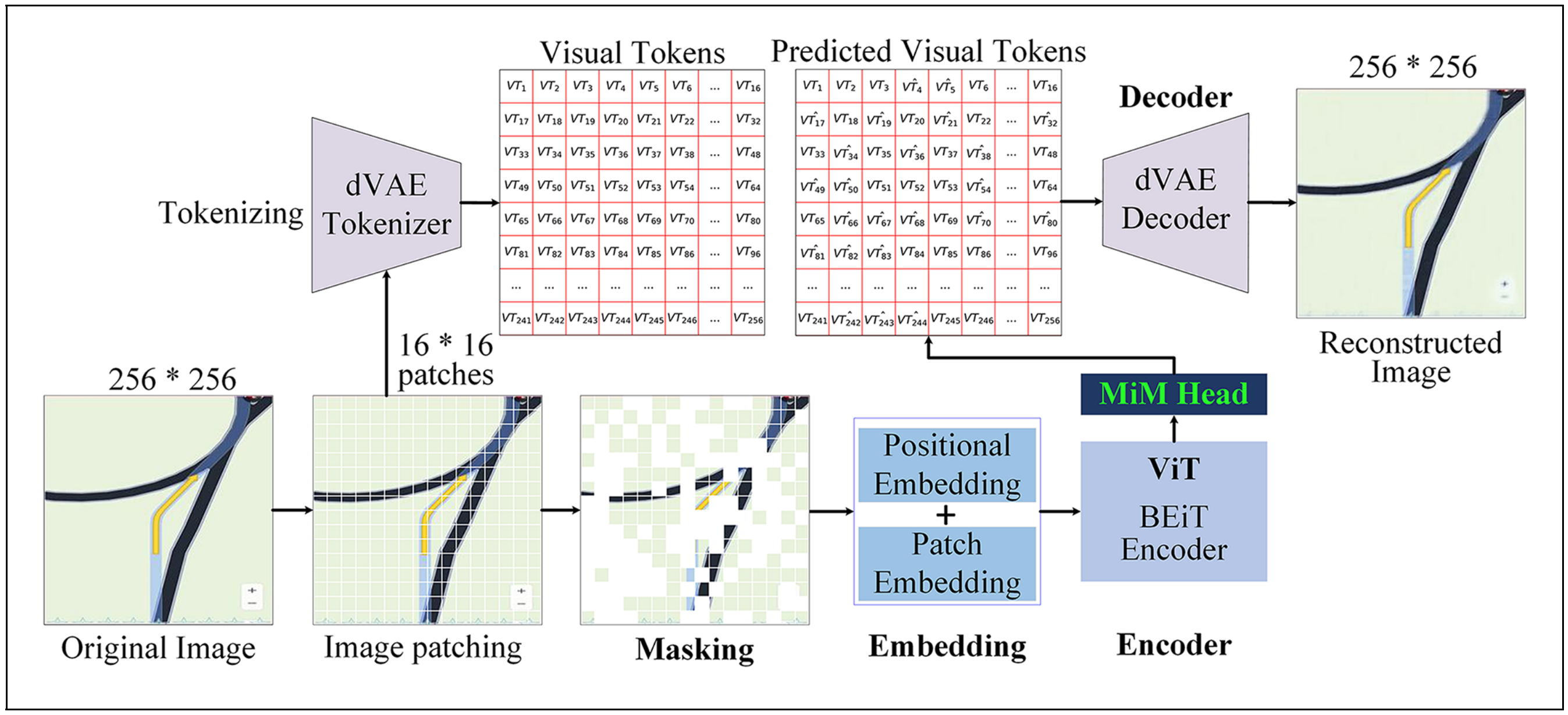

**Figure 4.** BEiT method pipeline for masked image modeling.
*Note*: dVAE = discrete variational autoencoder; BEiT = Bidirectional Encoder representation from Image Transformer; ViT = Vision Transformer.

The described MiM task, implemented through either the Uniform Masking method or the BEiT method, forces the model to learn meaningful representations of images by understanding the context of the unmasked patches. For the Uniform Masking method, the Swin Transformer encoder is pretrained using masked image regions, encouraging the model to effectively capture spatial relationships and hierarchical features. During the downstream classification task, the weights of the pretrained Swin Transformer encoder are retained, and the MAE Decoder is replaced by a classification decoder. In contrast, for the BEiT method, the ViT encoder is pretrained to predict discrete visual tokens corresponding to masked image regions. This approach emphasizes token-based representations that align with concepts in the visual vocabulary. For the classification task, the pretrained weights of the ViT encoder are preserved, and the MiM head is substituted with a task-specific classification decoder. Both methods leverage the robust features learned during the MiM task to enhance performance in the downstream tasks (i.e., the classification task of image types in this study), effectively transferring knowledge from the self-supervised pretraining phase to supervised fine-tuning.

This study also implemented and trained a ViT model without the proposed self-supervised pretraining as a baseline.

### *Customized Fine-Tuning*

In this paper, the lane-rendering images anomaly detection task is transferred into a two-, eight-, or nine-class (multi-label) classification problem, with the objective being to separate the seven types of anomaly from the normal images. The pretraining model weights in the self-supervised pretraining phase are transferred and further updated using the back-propagation mechanism with label smoothing cross-entropy as the loss function. To further boost the model performance, the MixUp technique (*44*) is adopted.

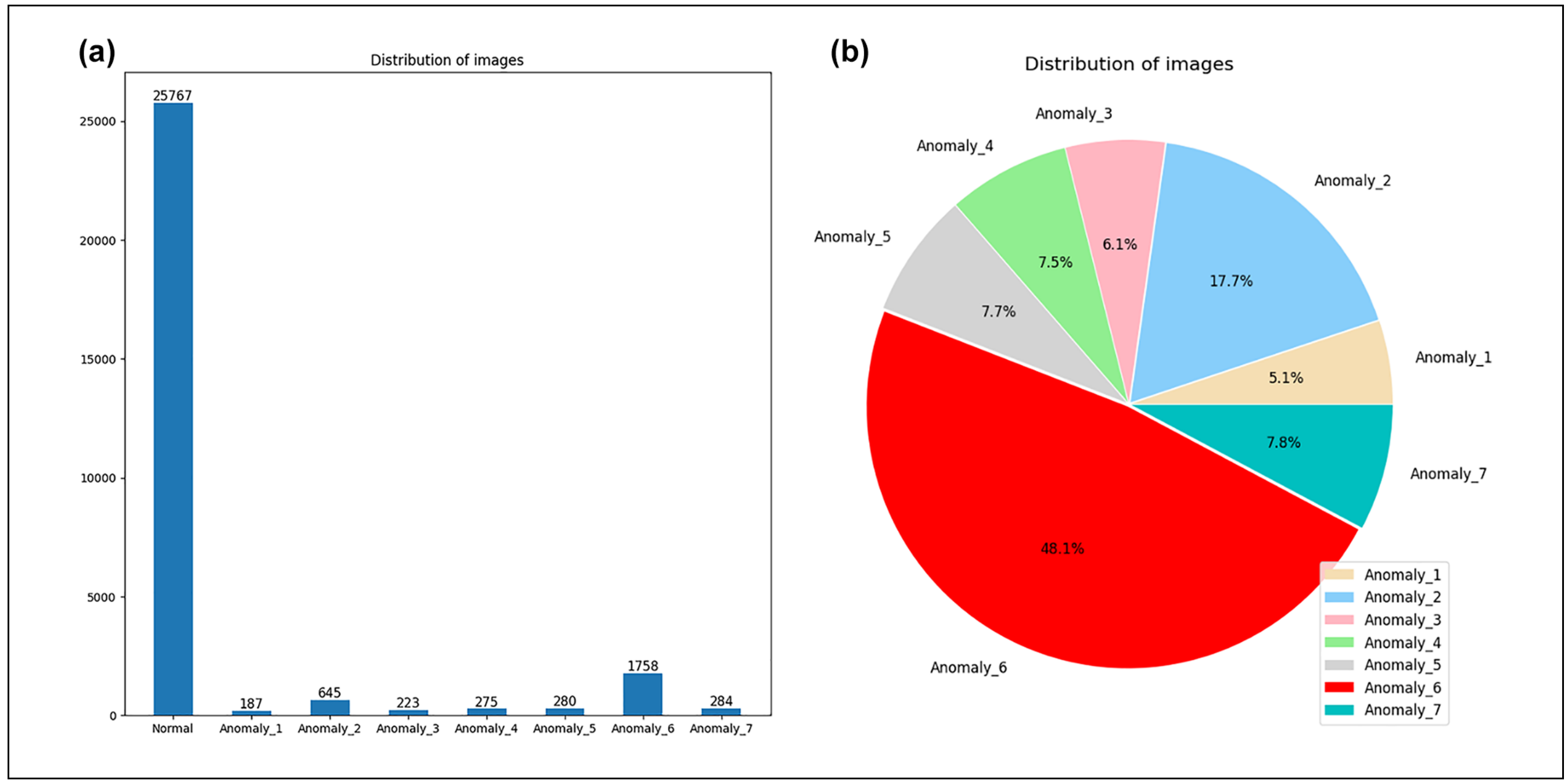

**Figure 5.** Distribution of labeled images. (*a*) Histogram plot for the distribution of all labeled images. (*b*) Pie chart for the distribution of each anomaly type within the labeled abnormal images.

### *Post-Processing*

After customized fine-tuning, during the testing stage, the fine-tuned model will be applied to assign "new" testing images that are unseen in the training process into the normal class or abnormal class. A post-processing phase is designed to aggregate the probability results and output the detection classification results.

In the post-processing, the neural network model outputs are first transformed into probabilities using the $softmax(\cdot)$ function, and then the probability of each image being abnormal is calculated and truncated/clipped with up and down thresholds. After obtaining the truncated probability, the final detection result can be determined by fine-tuning a probability threshold to distinguish the anomalies and the normal image samples.

## Experiments and Results

To verify the effectiveness of the proposed pipeline, extensive experiments were carried out under various settings.

### *Dataset Description*

The lane-rendering digital map image data used in this study are adjusted and rearranged from the 2022 Global AI Challenge. As mentioned, there are seven types of anomaly: Anomaly_1: The road center line extends out of the junction; Anomaly_2: The stop line is in the middle of a road; Anomaly_3: The navigation route does not match actual roads; Anomaly_4: The road shoulder is bumpy; Anomaly_5: A part of the road is missing; Anomaly_6: The road marking arrows overlap; and Anomaly_7: The lane lines overlap. Examples are shown in Figure 1.

In total, there are 161,772 images, with only 29,164 images labeled with the ground truth. Within the labeled ones, there are a total of 25,767 normal images and 3,397 images containing different kinds of abnormalities (note some images exhibit multiple different types of anomaly). Figure 5*a* shows the histogram plot for the distribution of all labeled images; Figure 5*b* illustrates the pie chart for the distribution of each anomaly type within the labeled abnormal images. It is visible and clearly observed that within the 29,164 labeled images, the majority are normal images. Furthermore, as illustrated in Figure 5, certain types of anomaly (e.g., Anomaly_6 and Anomaly_2) account for more samples than the other types of anomaly. Typically, Anomaly_6 accounts for nearly half (48.1%) of the total number of abnormal images.

The labeled dataset was randomly split into the training set, validation set, and test set at ratios of 70%, 15%, and 15%, respectively. The images were classified according to error types, and images with multiple error types were put into multiple categories. Thus, it is a multi-class multi-label classification problem, and there are a few more training examples than the image quantity. To be specific, in practice, the number of instances in the training set is 20,764, the number of instances in the validation set is 4,310, and the number of instances in the test set is 4,346. However, all of the available 161,772 images, regardless of whether labeled or not, are adopted in the self-supervised pretraining process.

### Tested Transformer Models

Two Transformer models (i.e., ViT [*40*] and Swin Transformer [*37*]) were implemented and tested in this study. The two Transformer models were tested both with and without self-supervised pretraining. Therefore, in total, there were four model variants: (1) pure ViT without pretraining, (2) ViT variant, BEiT, with the pretraining method described in Bao et al. (*25*), (3) pure Swin Transformer (Swin-Trans), and (4) Swin Transformer with Uniform Masking as self-supervised pretraining method (Swin-Trans-UM). The detailed model architectures (i.e., parameter settings for each layer of the tested models) are illustrated in Tables A1–A4 of the Appendix.

### Evaluation Metrics

Various metrics were used to evaluate the overall performance of the selected models. Four basic terms were first obtained: true positive (TP), which represents the number of correctly detected lane-rendering image anomalies; true negative (TN), which represents the number of correctly detected normal lane-rendering images; false positive (FP), which represents the number of incorrectly detected anomalies; and false negative (FN), which represents the number of incorrectly detected normal lane-rendering images. Then, based on the four basic metrics, accuracy, precision, and recall were calculated.

Accuracy is the percentage of correctly predicted lane-rendering image samples in regard to the total sample size, which can be defined as the following equation:

$$\text{Accuracy} = \frac{\text{TP} + \text{TN}}{\text{TP} + \text{TN} + \text{FP} + \text{FN}}. \tag{2}$$

Precision is the number of correctly predicted positive lane-rendering image anomalies as a percentage of the total number of predicted positive anomaly observations, and it shows how close the measurements are to each other. The mathematical expression of precision is defined by:

$$\text{Precision} = \frac{\text{TP}}{\text{TP} + \text{FP}}. \tag{3}$$

Recall ratio is the percentage of positive anomaly observations correctly predicted in the actual category:

$$\text{Recall} = \frac{\text{TP}}{\text{TP} + \text{FN}}. \tag{4}$$

The F1 score (F1) provides an overall view of recall and precision (weighted average). F1 ranges from 0.0 to 1.0, with 1.0 indicating perfect precision and recall. F1 can be obtained using the following equation:

$$\text{F1} = 2 \times \frac{\text{Precision} \times \text{Recall}}{\text{Precision} + \text{Recall}}. \tag{5}$$

Another appropriate indicator for evaluating the two-class classification problem is the receiver operating characteristic AUC. AUC assesses the model's ability to distinguish between normal and anomalous instances. It provides a single scalar value summarizing the trade-off between the TP rate (TPR) and the FP rate (FPR) across different thresholds, offering insights into the model's classification performance regardless of the specific threshold applied. Given its threshold-independent nature and its ability to encapsulate the model's discriminative power, AUC is particularly suitable for imbalanced classification problems, such as the lane-rendering image anomaly detection studied in this paper. Accordingly, this study selects AUC as the primary evaluation metric for comparing and assessing the performance of the tested models.

To measure AUC, one needs the TPR (i.e., recall ratio) and the FPR. The TPR and FPR can be obtained using the following two equations:

$$\text{TPR} = \frac{\text{TP}}{\text{TP} + \text{FN}} \tag{6}$$

$$\text{FPR} = \frac{\text{FP}}{\text{TN} + \text{FP}}. \tag{7}$$

### Experiment Setup

*Configuration Details.* In this paper, to reduce the computational payload and save training time, the size of the images for both the training set and test set was set to a resolution of 256 × 256. In pretraining, the proportion of masked patches was set to 75%. Experiments were carried out on four NVIDIA Tesla V100 (32 GB memory) GPUs, using PyTorch version 1.9.0 with CUDA Deep Neural Network library (cuDNN) version 11.1. The batch size was set to be as large as possible (i.e., 60). The learning rate was initially set to 0.001 with decay applied after each epoch.

*Data Augmentation.* A data augmentation technique, MixUp (*44*), where two samples (inputs and their labels)

are linearly combined, was adopted to upgrade the model performance. The idea of MixUp is to create new synthetic samples to encourage the model to make predictions based on more diverse data.

The new synthetic training sample $(\tilde{x}, \tilde{y})$ is given by:

$$\tilde{x} = \lambda x_a + (1 - \lambda) x_b, \tilde{y} = \lambda y_a + (1 - \lambda) y_b \quad (8)$$

where $x_a, x_b$ are two raw input sample vectors; $y_a, y_b$ are the corresponding one-hot encoded labels, and $\lambda$ is the MixUp parameter.

The MixUp technique helps the model generalize better by exposing it to more interpolated data points, leading to smoother decision boundaries.

*Loss Function Details.* As mentioned before, to make the proposed four-phase pipeline work, different loss functions were adopted accordingly in the pretraining and fine-tuning phases. In the pretraining phase, the MSE was selected as the loss function for the Uniform Masking method, since its objective is to reconstruct the masked patches directly at the pixel level. The cross-entropy loss function was employed for the BEiT method, since its MiM task involves identifying the correct visual token indices, framing the problem as a classification task over a visual vocabulary.

In the fine-tuning phase, the objective was to classify the lane-rendering images into normal ones and anomalies, which can be regarded as a typical classification task. The cross-entropy loss with label smoothing was adopted for this imbalanced classification task, which is illustrated in Equation 9:

$$\ell_{CE} = \ell(\boldsymbol{y}, \hat{\boldsymbol{y}}) = -(1 - \varepsilon) \log\left(\hat{y}_y\right) - \frac{\varepsilon}{C - 1} \sum_{c \neq y} \log(\hat{y}_c) \quad (9)$$

where $C$ is the number of classes; $\boldsymbol{y}$ is one-hot encoded true label vector; $\hat{\boldsymbol{y}}$ is the predicted probabilities output by the model over the $C$ classes—for example, $\hat{y}_y$ is the predicted probability for the true class and $\hat{y}_c$ is the predicted probability for any other class $c$; and $\varepsilon$ is the smoothing factor controlling the amount of uncertainty applied, usually set between 0 and 1.

With label smoothing, the true labels are adjusted to distribute some of the target probability mass to other classes. The overall effect of this modification is to provide a softer target. The model is less confident solely on one class, promoting better learning from non-ideal scenarios, such as label noise or ambiguity, and potentially improving generalization.

*Optimizer Details.* To efficiently train and validate the proposed model pipeline, different optimizers were tested in different stages. Four optimizers (Stochastic Gradient Descent, Adaptive Moment Estimation [Adam], Rectified Adam, and Adam with decoupled weight decay [AdamW] [*45*]), were tested in the pretraining and fine-tuning segmentation phases. Through the tests, AdamW performed the best in both the pretraining and the fine-tuning phases. Therefore, it was chosen for both of the phases.

For other hyperparameters and experiment implementations, this study generally followed the fine-tuned settings reported in Bao et al. (*25*), He et al. (*27*), and Li et al. (*39*).

### *Results*

Various experiments were carried out to compare the model performance of the tested four transformer models: pure ViT, pure Swin-Trans, BEiT, and Swin-Trans-UM. The obtained results of treating the problem as an eight-class classification task are illustrated in Figure 6 and Table 1.

From Table 1, it is evident that the significant differences in the number of fine-tuning epochs stem from the influence of the adopted MiM pretraining. The stopping criterion utilized in this study is AUC convergence. Specifically, fine-tuning is terminated when the improvement in AUC between consecutive evaluation epochs falls below a predefined threshold, signaling that the model's performance has stabilized.

With MiM pretraining, the Swin-Trans-UM and BEiT models converge in 15 and 41 epochs, respectively. In contrast, without MiM pretraining, the original Vanilla ViT requires 40 epochs, and the original Vanilla Swin Transformer demands 280 epochs to converge.

The adoption of MiM pretraining considerably reduces the total number of fine-tuning epochs needed for convergence. This is achieved by equipping the model with rich, context-aware semantic features during pretraining, which provide a robust initialization for the downstream classification task. As a result, models with MiM pretraining not only converge faster but also maintain or improve their classification accuracy. This observed disparity underscores the efficiency and effectiveness of MiM pretraining in lowering computational requirements while delivering high performance.

Furthermore, with regard to the primary and the most suitable overall model performance evaluation metric, AUC, both BEiT and Swin-Trans-UM outperform their variants without self-supervised pretraining. In particular, among the four models, Swin-Trans-UM obtains the best performance with regard to accuracy (94.77%), AUC (0.9743), recall (0.8022), and F1 (0.7805).

## Ablation Study

It is easy to identify that the quantity of abnormal and normal image samples is highly imbalanced. To alleviate

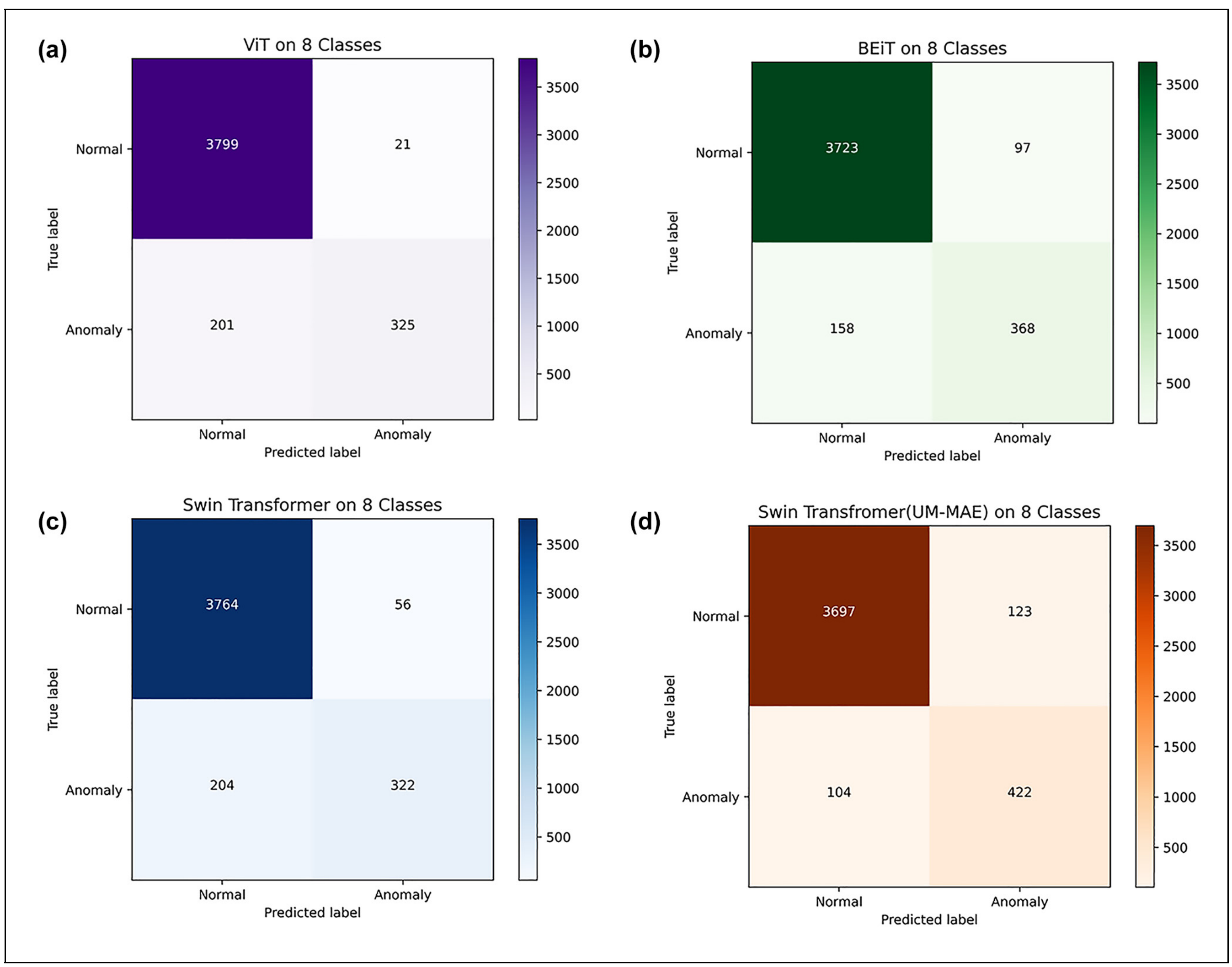

**Figure 6.** Testing results of the models visualized in confusion matrices. (*a*) Vision Transformer. (*b*) Bidirectional Encoder representation from Image Transformer. (*c*) Swin Transformer. (*d*) Swin Transformer with Uniform Masking as self-supervised pretraining method.

**Table 1.** Model Performance with Regard to Different Metrics

| Model | Accuracy | AUC | Precision | Recall | F1 score | Parameter (M) | Epoch time (s) | Number of fine-tuning epoch |
|---|---|---|---|---|---|---|---|---|
| ViT | 0.9489 | 0.9080 | **0.9393** | 0.6178 | 0.7454 | 632.20 | 4,210 | 40 |
| BEiT | 0.9413 | 0.9481 | 0.7913 | 0.6996 | 0.7427 | 311.53 | 159 | 15 |
| Swin-Trans | 0.9401 | 0.9498 | 0.8518 | 0.6121 | 0.7123 | 86.90 | 120 | 280 |
| Swin-Trans-UM | **0.9477** | **0.9743** | 0.7743 | **0.8022** | **0.7805** | 194.95 | 223 | 41 |

*Note*: AUC = area under the curve; ViT = Vision Transformer; BEiT = Bidirectional Encoder representation from Image Transformer; Swin-Trans = pure Swin Transformer; Swin-Trans-UM = Swin Transformer with Uniform Masking as self-supervised pretraining method.
Bold values in the table indicate the best performance for each metric.

this imbalance, two ablation studies were carried out using the Swin-Trans-UM model, with regard to the abnormal lane-rendering detection, not as the original eight-class multi-label classification problem but as a two-class classification problem (Swin-Trans-UM_2 as the corresponding model) or nine-class multi-label

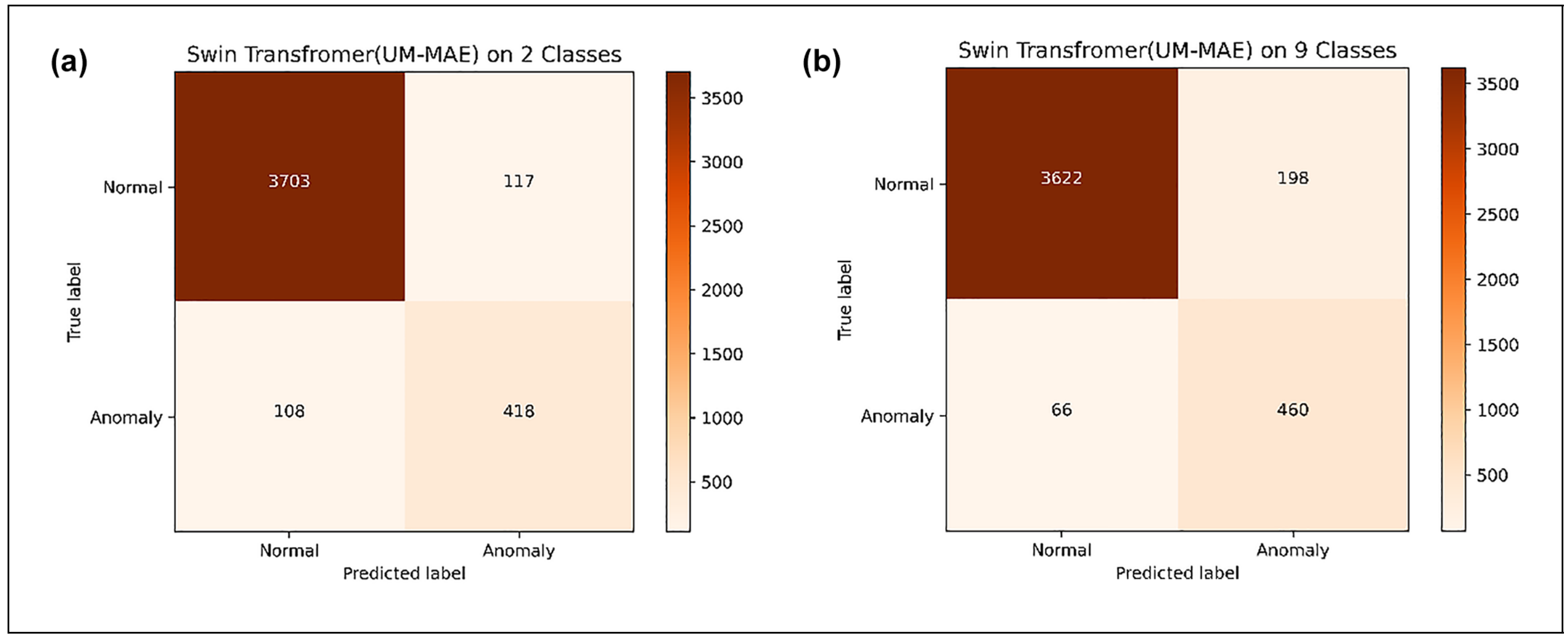

**Figure 7.** Confusion matrix of Swin Transformer with Uniform Masking as self-supervised pretraining method (Swin-Trans-UM) when treated as a two-class classification and a nine-class multi-label classification. (*a*) Swin-Trans-UM_2. (*b*) Swin-Trans-UM_9.

**Table 2.** Performance of the Swin-Trans-UM_2 and Swin-Trans-UM_9

| Model | Accuracy | AUC | Precision | Recall | F1 score |
|---|---|---|---|---|---|
| Swin-Trans-UM_2 | **0.9482** | **0.9756** | **0.7813** | 0.7947 | **0.7879** |
| Swin-Trans-UM_9 | 0.9392 | 0.9731 | 0.6990 | **0.8745** | 0.7770 |
| Swin-Trans-UM_8 | 0.9477 | 0.9743 | 0.7743 | 0.8022 | 0.7805 |

*Note*: AUC = area under the curve; Swin-Trans-UM = Swin Transformer with Uniform Masking as self-supervised pretraining method. Bold values in the table indicate the best performance for each metric.

classification problem (Swin-Trans-UM_9 as the corresponding model) in the fine-tuning process.

### *Treated as a Two-Class Classification*

When treated as a two-class image classification problem, all abnormal images are grouped as one class, and together with the normal class, there are two classes in the fine-tuning process. In this way, the imbalance between the classes is alleviated, since grouping abnormal classes together reduces the disparity between the number of normal instances and anomalies. By consolidating the abnormal classes into a single group, the number of anomaly-related instances is less sparse, making the distribution more balanced compared with treating each anomaly type separately.

The results of the tested Swin-Trans-UM_2 model performance under this setting are demonstrated in Figure 7*a* and Table 2. It is evident that, except for recall, all the other reported evaluation metrics (i.e., accuracy, AUC, precision, F1) for Swin-Trans-UM_2 are improved compared with the original approach, which treats the problem as an eight-class classification (Swin-Trans-UM_8).

### *Treated as a Nine-Class Multi-Label Classification*

When treated as a nine-class multi-label image classification problem, all abnormal images are grouped as one extra integrated class while keeping each sub-abnormal class as in the dataset. Thus, nine classes are obtained, and each abnormal instance will get at least two class labels. In this way, the imbalance between the classes is further alleviated. The results of the tested Swin-Trans-UM_9 model performance under this setting are demonstrated in Figure 7*b* and Table 2. Except for recall, all the other evaluation metrics of Swin-Trans-UM_9 are degraded compared with the original approach treated as an eight-class classification problem (Swin-Trans-UM_8). This might be because of the extra label for each abnormal instance confusing the model during the fine-tuning process when updating the model weights by backpropagation. Detailed explanations for this require further study.

## Conclusions, Limitations, and Future Research

Lane rendering is an important element in digital maps used for navigation services and other traffic-related applications. However, there might be anomalies in the lane-rendering images. To accurately and effectively detect the anomalies, this paper converts the problem of lane-rendering image anomaly detection to a classification problem, which allows various state-of-the-art computer vision techniques to be applicable. Furthermore, this paper proposes a four-phase pipeline, consisting of data preprocessing, self-supervised pretraining with the MiM method, customized fine-tuning using cross-entropy loss with label smoothing, and post-processing. Various metrics are adopted to evaluate the model performance. Extensive experiments have demonstrated that the proposed pipeline effectively addresses the lane-rendering image anomaly detection task, achieving outstanding performance with regard to high accuracy, F1, and AUC. In particular, self-supervised pretraining with MiM can greatly improve the model accuracy. For example, Swin-Trans-UM obtained better accuracy (94.77%) and better AUC (0.9743) compared with Swin-Trans, whose accuracy was 94.01% and AUC was 0.9498, while significantly reducing the model fine-tuning time. For example, Swin-Trans-UM reduced the number of epochs of Swin-Trans at 280 to only 41. Ablation study with regard to techniques to alleviate the data imbalance between normal and abnormal instances further enhances the model performance, with the two-class classification variant of the Swin-Trans-UM model—that is, Swin-Trans-UM_2—obtaining the best performance on almost all the evaluation metrics (i.e., accuracy [94.82%], AUC [0.9756], precision [0.7813], and F1 [0.7879]). Lastly, with regard to societal benefits, the proposed method can improve the efficiency of lane-rendering image data anomaly detection, reducing labor costs while maintaining high accuracy.

As for limitations, because of the unavailability of other relevant datasets, this study only examined and evaluated the proposed method and results on a single dataset, which might potentially constrain the generalizability of the proposed method and corresponding results. Furthermore, limited by the properties of the data, the focus of this study is confined to discerning whether the lane-rendering image is abnormal or normal. Further investigation into checking and diagnosing the specific anomaly types, as well as locating the anomalies within the images, could be intriguing directions for future studies. This would involve more detailed anomaly segmentation, which could provide valuable deeper insights into the nature and causes of detected abnormalities. However, achieving such advancements would require access to structured datasets equipped with labeled segmentation maps to facilitate robust anomaly localization and classification tasks.

Moreover, certain anomaly images in the dataset have multiple labels—a complexity that this study did not address. Future studies should explore methods for handling multi-label classification to account for overlapping or co-occurring anomalies. Techniques such as multi-label learning algorithms (*46*), label correlation modeling (*47*, *48*), or hierarchical classification approaches (*49*) could be explored to tackle this issue. Addressing multi-label scenarios would enhance the robustness and applicability of anomaly detection systems in real-world contexts.

Lastly, the current study employs a supervised approach during the fine-tuning phase, necessitating high-quality ground-truth labels. Future studies could explore the potential of semi-supervised or unsupervised machine learning approaches to distinguish anomalies from normal instances without relying on extensive labeled data. For example, Contrastive Language-Image Pre-training (CLIP) (*50*) can perform zero-shot classification by learning from large-scale, unannotated data, aligning images with textual descriptions. Similarly, Bootstrapping Language-Image Pre-training (BLIP) (*51*) can effectively perform image–text matching tasks in a self-supervised manner, which could help classify anomalies with minimal reliance on labeled data.

### Acknowledgments

The authors thank the 2022 Global AI Challenge for providing the original data.

### Author Contributions

The authors confirm contribution to the paper as follows: study conception and design: Y. Dong, X. Lu; data collection: Y. Dong, X. Lu, and R. Li; analysis and interpretation of results: Y. Dong, X. Lu, R. Li, and H. Farah; draft manuscript preparation: Y. Dong, X. Lu, R. Li, W. Song, B. van Arem, and H. Farah. All authors reviewed the results and approved the final version of the manuscript.

### Declaration of Conflicting Interests

The author(s) declared no potential conflicts of interest with respect to the research, authorship, and/or publication of this article.

### Funding

The author(s) disclosed receipt of the following financial support for the research, authorship, and/or publication of this article: This work was supported by the Applied and Technical Sciences (TTW), a subdomain of the Dutch Institute for Scientific Research (NWO) through the project “Safe and Efficient Operation of Automated and Human-Driven Vehicles in Mixed Traffic" (SAMEN) under contract 17187.

## ORCID iDs

Yongqi Dong https://orcid.org/0000-0003-1159-9584
Xingmin Lu https://orcid.org/0000-0001-8642-2658
Ruohan Li https://orcid.org/0000-0001-8102-7376
Wei Song https://orcid.org/0000-0003-0649-8850
Bart van Arem https://orcid.org/0000-0001-8316-7794
Haneen Farah https://orcid.org/0000-0002-2919-0253

## Supplemental Material

Supplemental material for this article is available online.